\documentclass[biblatex]{lni}
\usepackage{booktabs}

\usepackage[]{blindtext}

\begin{document}
\title[Ein Kurztitel]{Towards Sustainable Workplace Mental Health: A Novel Approach to Early Intervention and Support}
\author[1]{David W. Vinson}{}{}
\author[2]{Mihael Arcan}{mihael@luahealth.io}{0000-0002-3116-621X}
\author[2]{David-Paul Niland}{david-paul@luahealth.io}{0009-0001-1297-6532}
\author[2]{Fionn Delahunty}{fionn@luahealth.io}{0000-0002-2185-924X}%
\affil[1]{This work was conducted when the author was working at IntouchCX\\Winnipeg\\Canada}
\affil[2]{Lua Health\\Galway\\Ireland}
\maketitle

\begin{abstract}
In the contemporary workplace landscape, employee well-being has emerged as a critical concern. The American Psychological Association's report from 2021 revealed that a staggering 71\% of employees experienced stress or tension at work, with profound implications for workplace attrition and absenteeism, attributing 61\% of attrition and 16\% of sick days to poor mental health. A profound challenge faced by employers is that employees often remain unaware of their mental health issues until they reach a crisis point, resulting in limited uptake of corporate well-being benefits, even among those who statistically require such support. The primary objective of this research is to showcase the implementation of a groundbreaking stress detection algorithm capable of providing real-time support preemptively before individuals recognize the need for help. This solution leverages automated chatbot technology, offering a simple and controllable means of reaching employees. Our study comprehensively examines the feasibility of integrating these innovations into practical learning applications within real-world contexts, seeking to identify efficacious strategies and optimal deployment scenarios. We introduce a chatbot-style system integrated into the broader employee experience platform, designed to detect and address stress in real time. This platform encompasses schedule management, event calendars, performance evaluation, and internal communication, all aimed at enhancing employee well-being. The proprietary stress detection algorithm operates by analyzing chat conversations between employees and team leaders, flagging chats indicative of stress for immediate action. Unlike existing solutions, our algorithm objectively measures mental health levels and provides personalized treatment suggestions in real-time, utilizing linguistic biomarkers in employees' daily communication to passively identify mental health issues and deterioration. The system proactively engages with individuals through chat, offering a hyper-personalized well-being assistant experience. It detects signs of stress, burnout, depression, and anxiety and provides early access to resources, significantly improving support effectiveness, with studies demonstrating a 22\% increase when assistance is provided early. This study demonstrates the importance of enhancing the workplace environment for employees, easing work-related stress, and promoting sustainable well-being and support for the mental health of employees.
\end{abstract}

\begin{keywords}
Sustainable Workforce \and Stress detectors \and Mental Health \and Well-Being \and Chatbots
\end{keywords}

\section{Introduction}

The global population is experiencing an unprecedented burden of stress, exacerbated by factors such as economic uncertainties. According to the American Psychological Association (APA), 71\% of employees felt tense or stressed at work in 2021, contributing to workplace attrition and sick days. The challenge for employers lies in employees often not recognizing their mental health issues until they reach a crisis point. Despite investments in corporate well-being benefits, uptake remains low, with only 2-4\% of affected individuals utilizing available support.

In response to this crisis, the health technology sector has seen substantial growth, with mental health apps and tech exhibiting a compound annual growth rate of nearly 20\%. However, only a fraction of these tools is backed by research and clinical insights. The staggering cost of lost productivity due to depression and anxiety, estimated at \$1 trillion annually, necessitates robust solutions for workplace mental health.

Many existing applications in the health technology space outsource well-being initiatives, potentially leading to mismatches between organizational and individual needs. Moreover, self-reporting bias in time-consuming questionnaires undermines the accuracy of assessing mental well-being. Existing survey tools, grounded in clinical assessments, often miss the mark when applied to industry well-being. The objective of the presented work is to implement a novel stress detection algorithm that provides real-time support before individuals self-report needing help. The focus is on leveraging existing resources through a chatbot system, offering a simple and controllable solution. The study aims to assess the feasibility of such technologies within a real-world context, particularly in practical learning applications, and to provide valuable insights into effective practices and strategic implementation.

A chatbot-style system was implemented within an employee experience platform, targeting customer service employees. The platform, incorporating functionalities such as schedule management and internal communication, integrates a proprietary stress detection algorithm. This algorithm analyzes incoming chats between employees and team leaders, flagging specific chats indicating stress. Detected stress triggers an automated message based on predefined constraints.

Unlike other solutions, the implemented algorithm objectively measures mental health levels, offering personalized treatment suggestions in real-time. By analyzing linguistic biomarkers in employees' communication, the system passively identifies mental health issues and proactively engages individuals through the chat system, providing a hyper-personalized well-being assistant experience. The study aims to evaluate the effectiveness of this approach in a real-life industry scenario, assess individual differences in stress, and measure engagement. The research underscores the significance of enhancing the working environment, reducing employee work stress, and fostering sustainable work practices and well-being.

\section{Related Work}

In their examination of workplace mental health, \cite{Dewa2012} emphasize the prevalence of mental disorders among working-aged individuals, with mental health issues ranking among the top 10 and often coexisting with physical conditions. The study underscores the significant impact of mental health-related work disability, including unemployment, presenteeism, and absences. A cost analysis reveals potential expenses, indicating a company with 1,000 employees might spend up to \$171,000 annually on musculoskeletal disorders and up to \$378,000 on mental or behavioral disorders. The authors stress the importance of evidence-based interventions, cost considerations, and shared societal responsibility, urging each sector to contribute solutions and evaluate interventions' impact on disability for informed decision-making. Additionally, \cite{ijerph20021192} advocates an integrated approach to workplace mental health, targeting the reduction of work-related risk factors and promotion of mental well-being through positive work aspects. This strategy aims for universal solutions, addressing organizational culture, stigma, and disclosure norms, with a focus on balancing individual and organizational interventions, overcoming implementation barriers, and ensuring equity for diverse worker groups. Similarly, \cite{Taubman2023} highlights shifting expectations in workplace mental health, stressing the economic consequences of mental disorders. Despite challenges, organizations are urged to deliver effective interventions, with estimates projecting potential losses of \$6 trillion by 2030. The study explores workplace mental health standards, interventions, and assessments, emphasizing the importance of addressing legal considerations to foster a healthier workplace and alleviate individual suffering while mitigating societal burdens associated with mental health disorders.

The study by \cite{10.1016/j.ipm.2022.103011} addresses stress detection from textual data, recognizing its significant societal and economic impacts requiring early detection. Introducing a method that combines lexicon-based features and distributional representations, the study presents a comprehensive framework for stress classification. Evaluated on three public English datasets using various machine learning models, the approach, particularly the combination of FastText embeddings and select lexicon-based features, demonstrates effective performance, contributing valuable insights to stress detection research. Similarly, \cite{10192829} focuses on individual stress detection, employing deep learning techniques like LSTM and 1D CNN to identify stress levels from social media posts, primarily tweets. Successfully demonstrating the approach's effectiveness with Kaggle Twitter datasets and recent tweets, the study suggests the broader applicability of deep learning models for detecting various mental health conditions. Ethical concerns related to privacy and consent in future research are emphasized. Addressing speech-based stress detection, \cite{10246098} highlights its potential as a vital tool for early recognition, particularly among women. The study uses specialized speech features, specifically MFCCs and TEO-MFCCs, as inputs for a CNN classifier. The findings emphasize the superior performance of MFCCs in accurately predicting stress levels in women's speech, establishing them as a suitable choice for stress detection in this demographic. In a simulated office environment with N=90 participants, \cite{NAEGELIN2023104299} presents a machine learning approach for stress detection using multimodal data. Features from mouse, keyboard, and heart rate variability data are extracted to predict perceived stress levels, valence, and arousal. SHAP value plots interpret feature contributions, indicating that gradient boosting models using mouse and keyboard data outperform heart rate variability-based models. These findings fill methodological gaps in automated stress detection with implications for real-time stress detection in office settings. Lastly, \cite{antiwork2023} examines the connection between challenging work conditions, work-related stress, and mental health issues such as anxiety and depression. Focusing on workplace-specific solutions, the authors use the r/antiwork subreddit as a proxy for workplace dissatisfaction, creating a dataset for antiwork sentiment detection. The model, employing a RoBERTa feature extractor and RNN, identifies antiwork sentiments and uncovers factors contributing to dissatisfaction, aiming to help businesses enhance worker satisfaction and happiness by addressing the root causes of employee discontent.

In addressing the underdiagnosis of Major Depressive Disorder (MDD), \cite{fionn_aics18} proposes a passive diagnostic system that integrates clinical psychology, machine learning, and conversational dialogue systems. Using sequence-to-sequence neural networks, a real-time dialogue system engages individuals, while specialized machine learning classifiers monitor conversations to predict critical depression symptoms. Evaluation results suggest potential advancements in human-like chatbots and depression identification. Despite acknowledging data representation limitations and a small sample size, the study points towards enhancing support for individuals with MDD through real-time communication tools. Similarly, \cite{delahunty_smm4h19} introduces a deep neural network to predict PHQ-4 scores (depression and anxiety levels) from written text. Leveraging the Universal Sentence Encoder and a deep learning transformer neural network, the model demonstrates efficacy in psychometric score prediction. Exploring application to social media data, the study incorporates psycholinguistic features and a multi-dimensional deep neural network, noting challenges related to domain-specificity and generalizability. Addressing MDD, \cite{DBLP:journals/braininf/MilintsevichSD23} employs natural language processing to create a neural classifier detecting depression from speech transcripts. Predicting individual depression symptoms, the study utilizes a symptom network analysis approach and achieves comparable results to state-of-the-art models in binary diagnosis and depression severity prediction. Meanwhile, \cite{DBLP:conf/emnlp/BhatHABL21} focuses on toxic workplace communication in emails, introducing ToxiScope, a taxonomy to detect and quantify toxic language patterns. Utilizing annotation tasks and machine learning models, the study reveals insights into implicit and explicit workplace toxicity. The research suggests refining detection methods and exploring correlations between toxicity, power dynamics, and biases in workplace communication for future directions.

\section{Methodology}

As the world becomes more globalised than ever, we leverage machine translation models to translate any human-to-human communication into English. This was done especially to anonymize any sensitive communication between employees.\footnote{Research subjects were IntouchCX employees.} Our current neural machine translation (NMT) system covers translations from Spanish and Tagalog\footnote{Tagalog language, member of the Central Philippine branch of the Austronesian (Malayo-Polynesian) language family and the base for Pilipino, an official language of the Philippines (\url{https://www.britannica.com/topic/Tagalog-language/})} into English. Within this work, we used OpenNMT \cite{klein_opennmt_2017}, a generic deep learning framework mainly specialised in sequence-to-sequence models covering a variety of tasks such as machine translation, summarisation, speech processing and question answering. The architecture of our multi-language model, which covers English, Tagalog and Spanish within the same model, includes a 6-layer transformer with an attention mechanism for both the encoder and decoder. We trained the model for 50K steps on the training dataset and set the parameters of the model to the default configuration of OpenNMT. We used Sentencepiece \cite{DBLP:conf/emnlp/KudoR18} to split words into sub-word units.

\begin{figure*}[t]
    \centering
    \includegraphics[scale=0.55]{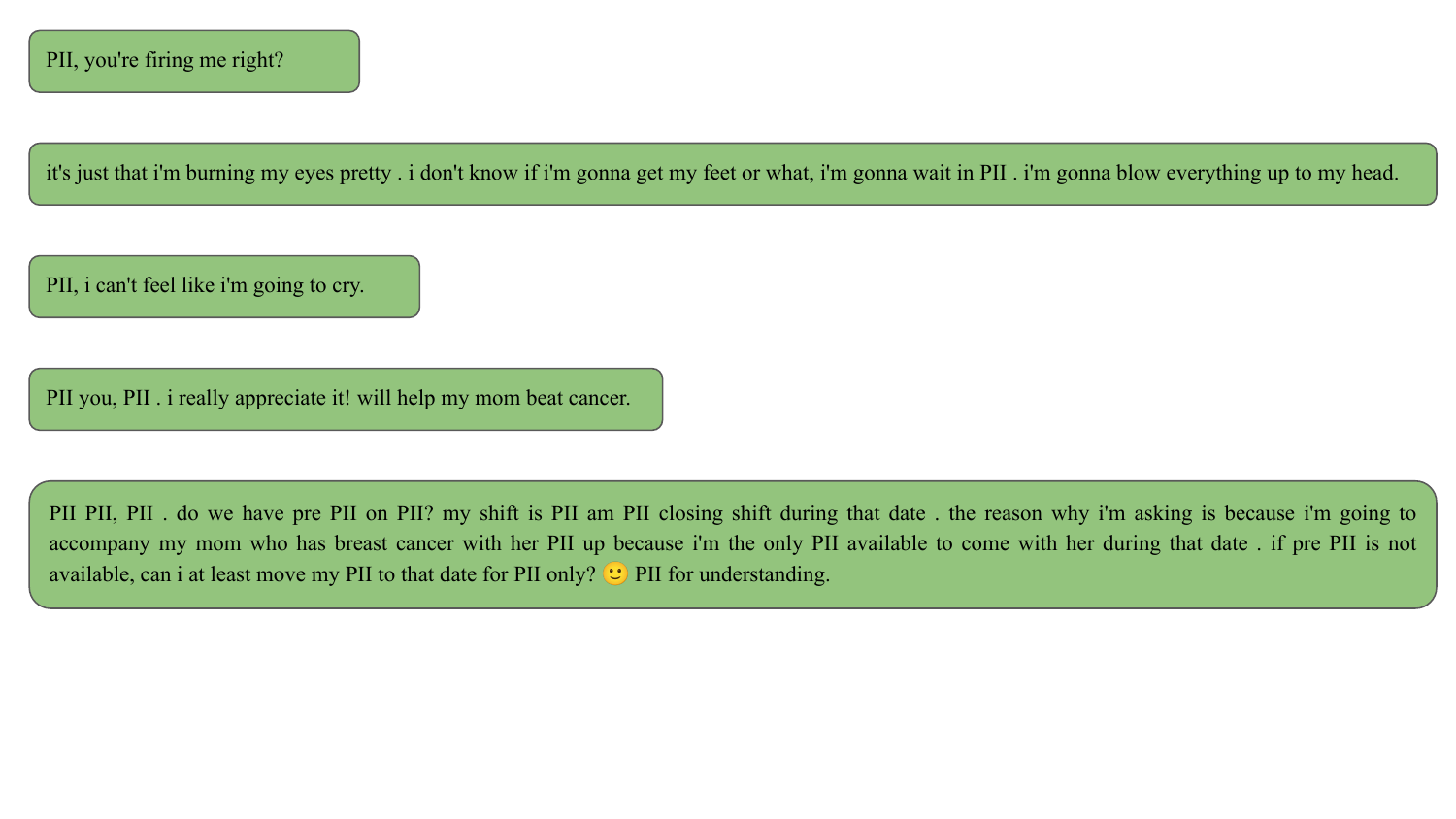}
    \vspace{-22mm}
    \caption{Text communication with redacted PII for confidentiality.}
    \label{image:chat}
\end{figure*}

To anonymize any personally identifiable information (PII) in the data stream, we leverage an ensemble of named entity recognition toolkits. For this, we use Spacy,\footnote{\url{https://spacy.io/}} which features a word embedding strategy using subword features and "Bloom" embeddings, a deep convolutional neural network with residual connections, and a transition-based approach. Secondly, we use Flair \cite{akbik2019flair}, which comprises popular and state-of-the-art word embeddings, such as GloVe, BERT, ELMo, and Character Embeddings, allowing us to combine different word embeddings. As the last resource, we leverage NLTK,\footnote{\url{https://www.nltk.org/}} which is built upon a tagger that labels each word in a sentence using the IOB format, where chunks are labelled by their appropriate type.\footnote{PII Classes identified: organisation, person, location, date, time, money, percent, facility, geo-political entity, numerical values, URLs, emails} Figure \ref{image:chat} illustrates the analyzed text with redacted PII for confidentiality.

Our algorithm can detect stress levels from written communication. This includes making use of linguistic biomarkers, which exhibit robust associations with a spectrum of mental health disorders \cite{doi:10.1177/0261927X09351676}.  This algorithm produces a binary output: a prediction of zero indicates the absence of detected stress, while a prediction of one signifies the presence of stress within a message. The algorithm's real-time analysis exclusively targeted internal communication messages exchanged among employees, excluding customer-facing interactions.

Upon analysis, an employee is notified if their messages flagged as stress exceeded five instances within the preceding seven-day period. This assessment occurred hourly and continued consistently throughout the day. Notably, a subsequent alert is withheld for a duration of four days following the issuance of an initial alert. This conservative approach was adopted to mitigate the risk of imposing additional stress upon the employee. Consequently, each employee could receive a maximum of two alerts within a given week.
]]\=v
Upon the initiation of an alert, an interactive chatbot is activated. The chatbot prompts a message, like, "I've noticed signs of potential stress. How are you feeling today?", whereby the employees were presented with a set of four radio button responses: [1, 2, 3, 4]. The ensuing chat session remains live, allowing employees to respond at their discretion. It is important to emphasize that the chat interaction can only commence upon activation by the chat system. This design choice was intentional, ensuring a transparent evaluation of the impact stemming from system-flagged messages and maintaining a clear demarcation from any gamification elements during the pilot phase.

Participants for this study were drawn from the employee cohort. A total of 416 employees were enrolled in the study, gaining access to the chatbot platform. These participants were geographically situated in Columbia and the Philippines. Due to data privacy and voluntary reporting, demographic information was only gathered from 377 participants. These participants fell within an age range of 19 to 57 years, with a mean age of 27.9 (SD = 7.13). The sample comprized 218 female and 159 male individuals; 181 were Columbia and 196 from the Philippines. 

Over a continuous span of 18 weeks from March 24th to July 28th during the year 2023, study participants were granted access to the chatbot platform. Throughout this duration, their internal communications with fellow colleagues were analysed via our machine-learning algorithms to detect indications of stress.

\section{Results}
In our study, we examined various topics to derive meaningful results. These topics encompassed Employee's Stress, Demographic Stress, Defining a 'Stressed' Cohort, Engagement, Engaged Cohort, Type of Engagement, and Depth of Engagement. Through the exploration of these subjects, we were able to uncover valuable insights that contributed to a better understanding of stress and engagement within our research context. Overall, there were a total of 210,314 messages that occurred within the 18-week pilot period, of which 1.99\% or 4,192 messages were flagged as stressful.

\begin{figure*}[t]
    \centering
    \includegraphics[scale=0.3]{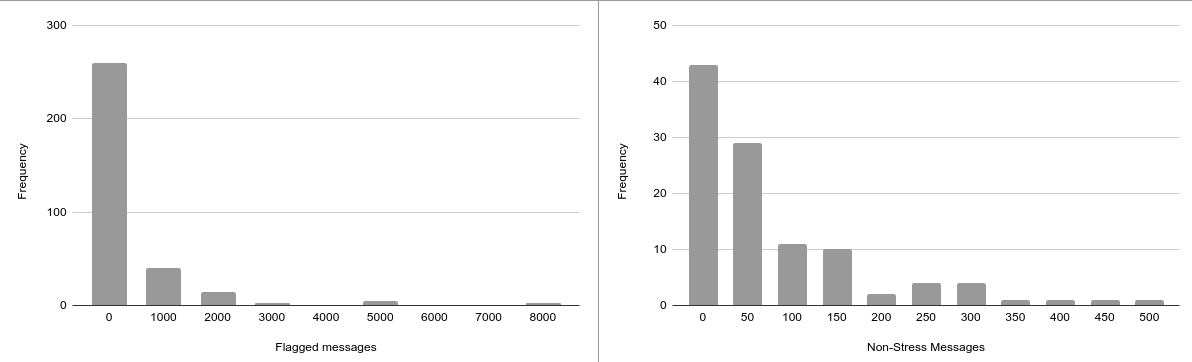}
    \caption{Histogram of the total number of messages per 309 employees who were flagged as having at least one stressed message over the 18 week pilot (left side). Histogram of the total number of messages per 107 employees who never had a stressed message over the 18 week pilot (right side).}
    \label{image:fig1}
\end{figure*}

\paragraph{\textbf{Employee’s Stress}} A total of 314 out of 416 employees were flagged as having at least one stressed message with a median of 363 messages per employee. On the other hand, 107 participant’s messages were never flagged by our algorithm with a median of 73 messages per employee. Employee message count was highly skewed (see Figure \ref{image:fig1}), while there was a highly significant correlation between total message count per employee and total flagged messages; r(414)=0.8 (t=27.2, p<0.001). With more messages there is a higher probability of the occurrence of a stressed message. However, the percentage of stressed messages does not correlate with the total number of stressed messages (r(414)=0.076). For this reason, we report the percentage of stressed messages in our analyses moving forward. 

\begin{figure*}[t]
    \centering
    \includegraphics[scale=0.5]{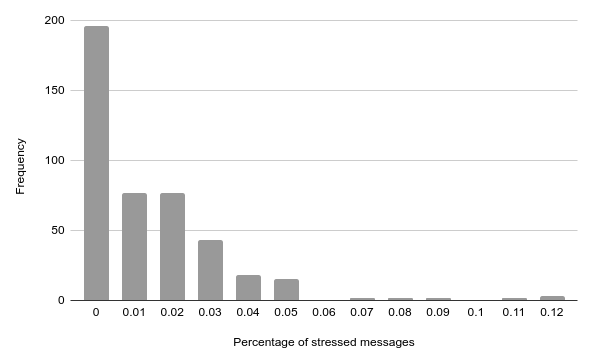}
    \caption{Histogram of percentage of stressed messages by employee.}
    \label{image:fig2}
\end{figure*}

The distribution for the percentage of stressed messages by employee was also highly skewed (Figure \ref{image:fig2}) with median values of 1.2\% stressed messages per employee (M = 1.6\%, SD = 1.79). While some employees experienced as much as 12.5\% stressed messages, such a skewed  representation indicates that there may be a stressed cohort which we discuss later on. 

\begin{figure*}[t]
    \centering
    \includegraphics[scale=0.5]{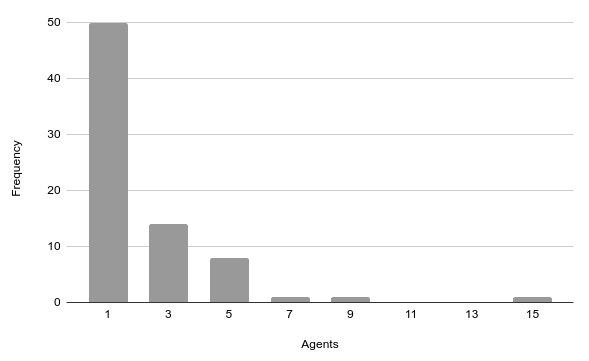}
    \caption{Distribution representation of repeated agent responses.}
    \label{image:fig5}
\end{figure*}

Furthermore, 1.84\% of messages were flagged as containing stress on average by day; with an average distribution of stress being normal. However variance in stress differed. 
There were only five days throughout the length of the pilot where stress was not detected in any messages. Three of the five days landed on weekends, where generally client activity is slower. The total message count (M=187.2) was roughly 10\% of the total average message count per day (M=1882.8). On average, 22 participants have at least one stressed message per day as having at least one stressed message per day (SD = 15.37) ranging from 1 to 62 messages. However, as can be seen in Figure \ref{image:fig5}, the distribution appears bimodal, such that many days resulted in less than 10 participants flagged as having stress in their message.

sample-base.bib
\paragraph{\textbf{Demographic Stress}} Furthermore, we were interested in determining if there were statistically significant differences among agent demographic variables and stress. Specifically, do certain demographic variables predict higher levels of stressed messages? To do this, we took the percent of stressed messages for each agent across the entire pilot. We matched this data with the available demographic data. This resulted in 369 agents. We constructed a linear model 
of stressed messages for each agent as a function of \texttt{Age}, \texttt{Gender}, \texttt{Location} and \texttt{Tenure}.  Based on the linear model, we learned that the model was significant (F(4,365) = 56.1, adjR2 =0.375, p<.001). Individual t-tests show significant values for \texttt{Tenure} (t = 3.13, p=.002) and \texttt{Location}, such as agents located in \texttt{Colombia} showed a significantly higher percentage of stressed messages compared to agents in the \texttt{Philippines} (t=14.00, p<.001). After analysis, we found that 94 participants who did not have any stressed messages. As such the results of our model may be zero inflated. For this reason, we ran two more models: a (\textit{i}) linear regression model specifically looking at the number of instances of stressed messages, adjusting the model to fit the skewed data seen by the representation of zeros in the data. In addition to that, we ran a (\textit{ii}) logistic regression, where we classified agents as either having had stress or no stress throughout the pilot. 

We ran linear regression with only those agents who had at least once instance of stress, which demonstrated  statistically significant results (F(4,270)=19.44, adjR2=0.21, p<.001). Individual t-tests show significant values for \texttt{Tenure} (t=2.69, p=0.008) and \texttt{Location}, such as \texttt{Colombia} showed a significantly higher percentage of stressed messages than the \texttt{Philippines} class (t=7.87, p<0.001). Interestingly, \texttt{Age} also was statistical significant, such that younger agents are more likely to have a higher percentage of stressed messages (t=-2.012, p<0.05). 

The logistic regression model shows statistically significant differences in \texttt{Location}, such that agents in \texttt{Colombia} were significantly more likely to have stressed messages. The log odds indicate that if an agent works in \texttt{Colombia} they are 4 times more likely to have at least one stressed message throughout the pilot compared to the Philippines (z = 6.619, p<.001, estimate = 3.99). Interestingly agent \texttt{Tenure} trended toward statistical significance (t = 1.73, p<0.1), such that for every additional week an employee stays with the organization the log odds of that agent having at least one reported instance of stress in their messages increased by 0.03. 

Additionally, we were interested to see if there was a higher likelihood of stressed messages occurring for specific demographics of our agents. Therefore, we leveraged the Negative binomial distribution model, which looks at the probability of count data. Our findings support our previous model findings that agents in Colombia are more likely to have stressed messages (estimate=0.75, p<0.001). As well as \texttt{Tenure}; for every unit increase in \texttt{Tenure}, you are 0.053 times more likely to have a stressed message (p<0.001). Interestingly, it also appears that your 0.34 time more likely to have a stressed message if you are male (p<0.05). Collectively, these model findings support the above described findings, suggesting a robust relationship between \texttt{Location} and \texttt{Tenure}. While additional studies and analyses beyond the scope of the current work would need to be conducted to assess the possible impact of \texttt{Gender} and \texttt{Age}. 

Our findings indicate that there is a significant difference in \texttt{Location} and \texttt{Tenure} such that those who work in Colombia are more likely to show stress in their messages compared to those who do not. In addition, those who have been with the organization for longer, are more likely to show stress in their messages as well. While this finding suggests that different groups may be more stressed than others, cultural differences regionally and within organizations need to be considered. For instance, employees in Colombia may be more likely to simply voice their concerns over other regions, while more tenured agents may feel safer in their jobs, thus more likely to also voice their concerns. There is a significant positive correlation between age and tenure (r=0.14, t=2.7, p <0.05). While the correlation is weak, it suggests that the possibility that younger employees voicing stress is not related to tenure. Again, this suggests that varying cultural differences, location, age, and organization tenure, may all influence whether an employee is likely to voice stress in their digital communications in the workplace.

\paragraph{\textbf{Defining a ‘Stressed’ Cohort}} We further aimed to determine if there were statistically significant differences between participants who fell above two standard deviations away from the average percentage of stress for all participants. The distribution of stressed messages was highly skewed with some agents showing up to 12.5\% of their messages containing stress indicators. To determine a cohort, we subset participants whose amount of stressed messages was more than two standard deviations above the mean, i.e., those who fall outside 95\% of the total distribution. Participants in this cohort have at least 5.18\% of their total messages flagged as containing stress. This resulted in a cohort of 15 individuals, whereby for three participants demographic data is missing. Of the remaining 12 agents, 91\% were from Colombia (11 out of 12). A t-test indicated a statistical significant difference in \text{Tenure} for the stressed cohort (M=18.2 months), compared to the overall sample (M=12.24, t=2.24, p=0.045). While there were no significant difference in \texttt{Age} (t=0.5, p>0.05, M=26.7) compared to our population (M=27.95) or \texttt{Gender}. 

Our findings indicate that there is a statistical difference such that agents are more likely to be flagged as having stress in their messages if they work in Colombia and have been with the organization longer.

\paragraph{\textbf{Engagement}} As we were also interested in tracking engagement, we outlined the constraints on chatbot-initiated conversations as well as other limitations.  
Within our framework, a chat was initiated, if the algorithm flagged more than five messages containing stress, within a seven-day period. Igniting a second chat was then blocked for five days, and an initial 48 hours from the last interaction with the chatbot on the agent side. As an example, if the chat was initiated on a Wednesday, and the agent interacted with it on a Thursday, 48 hours later is still within the five-day time frame. However, if an agent went on leave for two weeks and did not interact with the initial chat for two weeks, another 48-hour window was added. 


There were a total of 537 alerts sent to employees during the 18-week period, of which 185 lead to employee engagements, i.e., employees responded to the message. Interestingly, only 26\% of agents, (n=109) were ever alerted. This reflects the conservative nature of the alert system and the degree to which we chose not to reach out unless there were an abundance of stress indicators throughout a series of messages. The average alert rate for these agents was 5.37 times per participant across the entire pilot.

Of the 109 agents who were alerted 74 responded at least once. While the majority of the 109 employees responded twice to pings (median, SD = 2.27), reflecting the heavy tailed distribution in Figure \ref{image:fig5}. There was a strong correlation between response rate and ping rate (r=0.65, t=8.78, p<0.001), such that the more pings you received, the higher your response rate. This indicates that the more times an agent has an opportunity to engage, the more likely they are to engage with the system. 

\paragraph{\textbf{Engaged cohort}} The above findings allow us to determine users who were highly engaged, or more likely to initially engage with the chatbot system when flagged as being under stress. To determine a cohort that was highly engaged, we first removed all agents who were only pinged one time. We do this as the median average is two, and while some agents may respond 100\%, they may only have been alerted one time, possibly skewing the distribution. This dropped the total number of agents from 109 to 87. From here, we took a standard cutoff of 50\%. That is, we only look at those agents who were likely to engage with the system at least 50\% of the time when pinged. This resulted in 24 agents. 

Demographic data was available for 22 of the 24 individuals identified as part of the engaged cohort. The engaged cohort consisted of 11 males and 11 females with an average age 25.5, mostly residing in Colombia (n=17), compared to the Philippines (n = 5). Tenure of the engaged cohort was 11.16 months. Statistical analysis found differences in \texttt{Location} and \texttt{Age} compared to the sample population. Such that, the engaged cohort was younger (t=2.4, p<0.05, d.f.=26) and consisted of more individuals from Colombia ($\chi^2$=8.1, d.f.=1, p<0.001). No significant differences in \texttt{Gender} for the engaged group were found compared to the overall sample population. 

\begin{figure*}[]
    \centering
    \includegraphics[scale=0.4]{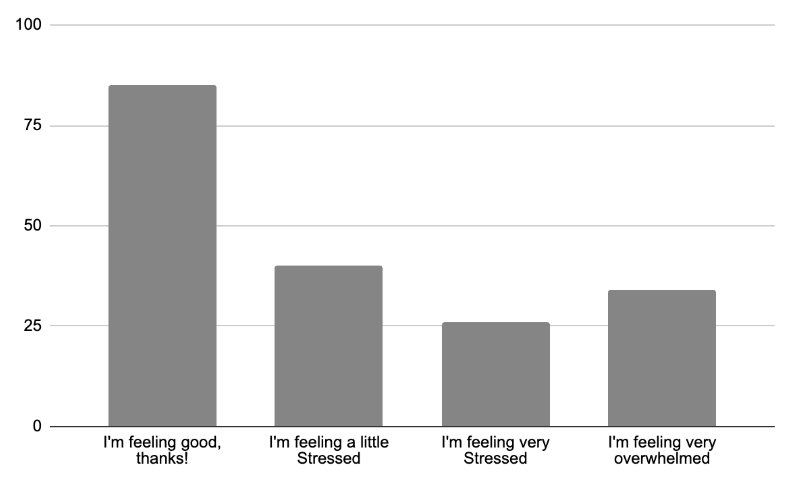}
    \caption{The total number of responses by category.}
    \label{image:fig7}
\end{figure*}

\paragraph{\textbf{Type of Engagement}} We further study the type of engagement, whereby agents responded a total of 184 times out of a total of 537. Figure \ref{image:fig7} below outlines the 4 types of responses as well as their total response count (and percent as a function of the total number of responses). Agents were more likely to respond \textit{I’m feeling good thanks} than any other category followed by \textit{I’m feeling a little stressed} and \textit{I’m feeling very stressed}. However, and interestingly, \textit{I’m feeling very overwhelmed} did not follow this downward trend. Instead the data’s U-shaped function indicates that agent’s self-reported stress may be bimodal such that while most agents self-report little to no stress, of those who do report higher levels of stress, they may be more likely to report being overwhelmed. 

In addition to understanding agent’s responses we were also interested in understanding if there were differences among the types of agents who responded. To do this, we ran an ordinal logistic regression model with response type as a factor of \texttt{Age}, \texttt{Gender}, \texttt{Location} and \texttt{Tenure}. We learned that employees from Colombia (\texttt{Location} class) were 2.36 times more likely to report higher levels of stress (t = 2.59, p<0.01). All other variables (\texttt{Age}, \texttt{Gender}, \texttt{Tenure}) did not significantly predict increases in self reported stress.

\begin{figure*}[]
    \centering
    \includegraphics[scale=0.5]{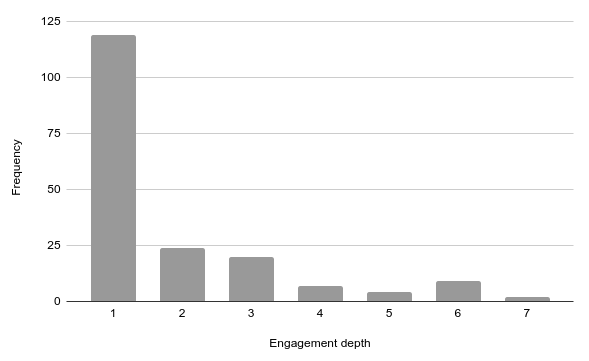}
    \caption{Histogram of the depth of engagement.}
    \label{image:fig6}
\end{figure*}

\paragraph{\textbf{Depth of Engagement}} We define depth as the number of back-and-forth interactions an agent had with the chat system. The average depth of agent interactions was M=1.76 (SD = 1.33). As it can be seen in Figure \ref{image:fig6}, agents' interaction depth is skewed such that many agents interacted with a depth equal to one, while a small set of agents interacted anywhere between two and four times.

A multiple regression model with depth as a factor of \texttt{Age}, \texttt{Gender}, \texttt{Location} and \texttt{Tenure} found that the depth of engagement differed on \texttt{Tenure} such that agents are less tenured are more likely to have longer interactions with the chat system (t=2.83, p<0.005, F(4,420)=3.29, adjR2=0.02, p<0.05). In addition, \texttt{Location} class was seen as trending toward significance, such that agents in Colombia were more likely to have a deeper engagement with the chat system than agents in the Philippines (t=1.66, p=0.097). When non-responses were removed, there were no significant differences among demographic variables of interest. 

Lastly, crucial to the adoption of any such system is the understanding of who is likely to interact. 
Given that, there were many instances where agents did not respond to prompts. We also ran a logistic regression between agent responses and non-responses. The purpose of this was to determine if certain types of agents were more likely to interact with the system than others. Results reflect the linear regression model above such that agents who were less tenured were more likely to respond in general (t=3.87, p<0.001) while agents from Colombia were also found to be trending toward significance (t=1.88, p=0.06). This again indicates that less tenured employees and those from Colombia were more likely to interact with the system from the very beginning compared to less tenured agents or those in the Philippines.

\paragraph{\textbf{Self-Reporting}} 
Similar to the above analysis on messages containing stress, we were also interested in understanding the cohort of employees who self-reported that they were \textit{Very overwhelmed}. When individuals selected this, they were pointed directly to EAP\footnote{EAP - Employee Assistance Programme} services and provided access to helplines as needed as well as region-specific well-being support. Because of this, tracing their depth of engagement and content accessed would likely be skewed. However, uncovering similarities among employees who self-report along the higher end of stress can provide critical information to programs in support of stressed populations. It is with this in mind that we aim to determine whether a self-reported stressed cohort can be statistically identified. 

There were a total of 34 instances where employees indicated they were very overwhelmed. However there were only 19 unique individuals who were alerted. One employee responded that they were very overwhelmed six times, three employees responded they were overwhelmed three times, three responded they were overwhelmed on two separate occasions and the remaining 12 responded they were overwhelmed on only one occasion. 

There was a weak significant difference in \texttt{Tenure} such that those who self-reported high stress (\texttt{Very overwhelmed}) have been with the organization longer (M=17.2 months) compared to the general sample population (M=12.17, t=1.97, p=0.06). There was no difference in age between the general population and self-reported high stress. Furthermore we observed that there was also a weakly significant difference in \texttt{Gender}, such that more males were likely to self-report high stress compared to the general population ($\chi^2$ = 3.612, p=0.057). Additionally, there was a significant difference between self–reported high stress and location, such that individuals from Colombia were more likely to self-report higher stress compared to the general sample population ($\chi^2$=10.5, p<0.005). Collectively these findings suggest that individuals were more likely to self-report high stress if they were from Colombia, had been with the organization longer and were also male.



\section{Conclusions}

In this scientific publication, a comprehensive analysis was conducted on stress-related messages within an 18-week pilot program, comprising 210,314 messages. Out of these, 1.99\% (4,192 messages) were identified as containing stress indicators. The study explored examples of these messages, revealing that 314 out of 419 employees had at least one stressed message, with a median of 363 messages per employee. Employee message counts exhibited a skewed distribution, with a significant correlation between total message count and the likelihood of encountering a stressed message.

Demographic factors such as Age, Gender, Location, and Tenure were assessed for their influence on stress levels. Location, specifically Colombia, and longer tenure were associated with higher percentages of stressed messages, as supported by logistic regression models.

The analysis identified a "stressed cohort," a subset of individuals with significantly higher percentages of stressed messages, mostly from Colombia and with longer tenures. Engagement with chatbot-initiated conversations was briefly examined, showing a positive correlation between engagement opportunities and actual engagement. Less tenured employees were more likely to have deeper interactions with the chat system.

The study emphasizes the potential impact of alert timing on employee engagement and well-being, emphasizing the intersection between sustainability and mental health in the workplace. By focusing on enhancing the working environment, mitigating work-related stress, and fostering sustainable work practices, it advocates a holistic approach to employee well-being. The findings underscore the connection between sustainability initiatives and mental health, offering valuable insights for organizations striving to create healthier and more resilient workplaces. Understanding productivity seasonality and employee stressors can inform the impacts of dynamic alert timing and tracking on well-being tool engagements. Future research aims to delve into the impact of timing on employee engagement by tracking the time between alerts and interactions.

\printbibliography

\end{document}